# Behavior Trees in Action: A Study of Robotics Applications

Razan Ghzouli
Chalmers | University of Gothenburg
Sweden

Thorsten Berger
Chalmers | University of Gothenburg
Sweden

Einar Broch Johnsen
University of Oslo
Norway

Swaib Dragule
Chalmers | University of Gothenburg
Sweden

Andrzej Wąsowski
IT University of Copenhagen
Denmark

## Abstract

Autonomous robots combine a variety of skills to form increasingly complex behaviors called missions. While the skills are often programmed at a relatively low level of abstraction, their coordination is architecturally separated and often expressed in higher-level languages or frameworks. Recently, the language of Behavior Trees gained attention among roboticists for this reason. Originally designed for computer games to model autonomous actors, Behavior Trees offer an extensible tree-based representation of missions. However, even though, several implementations of the language are in use, little is known about its usage and scope in the real world. How do behavior trees relate to traditional languages for describing behavior? How are behavior-tree concepts used in applications? What are the benefits of using them?

We present a study of the key language concepts in Behavior Trees and their use in real-world robotic applications. We identify behavior tree languages and compare their semantics to the most well-known behavior modeling languages: state and activity diagrams. We mine open source repositories for robotics applications that use the language and analyze this usage. We find that Behavior Trees are a pragmatic language, not fully specified, allowing projects to extend it even for just one model. Behavior trees clearly resemble the models-at-runtime paradigm. We contribute a dataset of real-world behavior models, hoping to inspire the community to use and further develop this language, associated tools, and analysis techniques.

## 1 Introduction

The robots are coming! They can perform tasks in environments that defy human presence, such as fire fighting in dangerous areas or disinfection in contaminated hospitals. Robots can handle increasingly difficult tasks, ranging from pick-and-place operations to complex services performed while navigating in dynamic environments. Robots combine skills to form complex behaviors, known as missions [25, 34]. While skills are typically programmed at a relatively low level of abstraction (such as controllers for sensors and actuators), the coordination of skills to form missions in higher-level representations is becoming increasingly important.

Behavior Trees are attracting attention of roboticists as a language for such high-level coordination. They were originally invented for computer games, to define the behavior of autonomous non-player characters. Similar to autonomous robots, non-player characters are reactive and make decisions in complex and unpredictable environments [29, 31]. Their popularity in robotics stems from their modularity and malleability when expanding or debugging missions [3, 11, 13–15, 23, 36, 37]. Users appreciate a purportedly easy-to-understand hierarchical structure, able to represent layers of behavior. Traditionally, missions have been specified using finite state machines, but the representation of complex and dynamic surroundings quickly makes state machines unmanageable [29]. Hierarchical state machines [28] overcame these issues, bringing modularity and the structuring of tasks into sub-tasks. Still, many find evolving hierarchical state machines harder than evolving behavior trees [7, 14, 35].

We present a study of behavior tree languages and their use in real-world robotic applications. Specifically, we ask:

**RQ1.** *What are the key characteristics, modeling concepts, and design principles underlying behavior tree languages?*

**RQ2.** *How are the language concepts used in robotic projects?*

**RQ3.** *What are characteristics of Behavior Trees models?*

To answer these questions, we mine open-source repositories for behavior trees in robotics applications and analyze their usage. Behavior tree implementations (i.e., libraries) are identified and analyzed as domain-specific languages (DSLs).

We find that Behavior Trees are a pragmatic language, not fully specified, allowing, even expecting, concrete projects to extend it by-need. The use of behavior trees in robotics follows the *models-at-runtime* paradigm [4, 5]. Models are used for coordinating skills, actions, and tasks, which are implemented by lower-level means (e.g., The Robot Operating System (ROS) components). We hope to raise the interest of the software languages and modeling research communities in behavior trees and their usage. We also hope that this analysis can inspire designers of behavior tree languages in robotics to revisit, or at least justify, some design choices. We contribute a dataset of real-world behavior models, hoping to inspire the community to use and further develop this language, associated tools, and analysis techniques.

An accompanying online appendix [1] contains the models dataset, mining and analysis scripts, and further details.

## 2 Background

Behavior trees are well-suited to express the runtime behavior of agents, which has fueled applications in computer games and robotics. High-profile games, such as Halo [31], use behavior trees. In the robotic community, there has been a growing interest in behavior trees. There was a dedicated workshop on behavior trees in robotics at IROS'19,[1] one of the key research conferences in robotics. ROS, the main open source platform for robotics software, has recently adopted behavior trees as the main customization mechanism for their navigation stack.[2] In addition, multiple projects in RobMoSys, one of the leading model-driven community in robotics,[3] have been launched to create a set of best practices and tools for behavior trees (e.g., CARVE[4] and MOOD2Be[5]). The EU project Co4Robots[6] developed a mission-specification DSL for multiple robots upon behavior tree concepts [23, 24].

[1] https://behavior-trees-iros-workshop.github.io/
[2] https://github.com/ros-planning/navigation2/tree/master/nav2_behavior_tree
[3] https://robmosys.eu/
[4] https://carve-robmosys.github.io/
[5] https://robmosys.eu/mood2be
[6] http://www.co4robots.eu

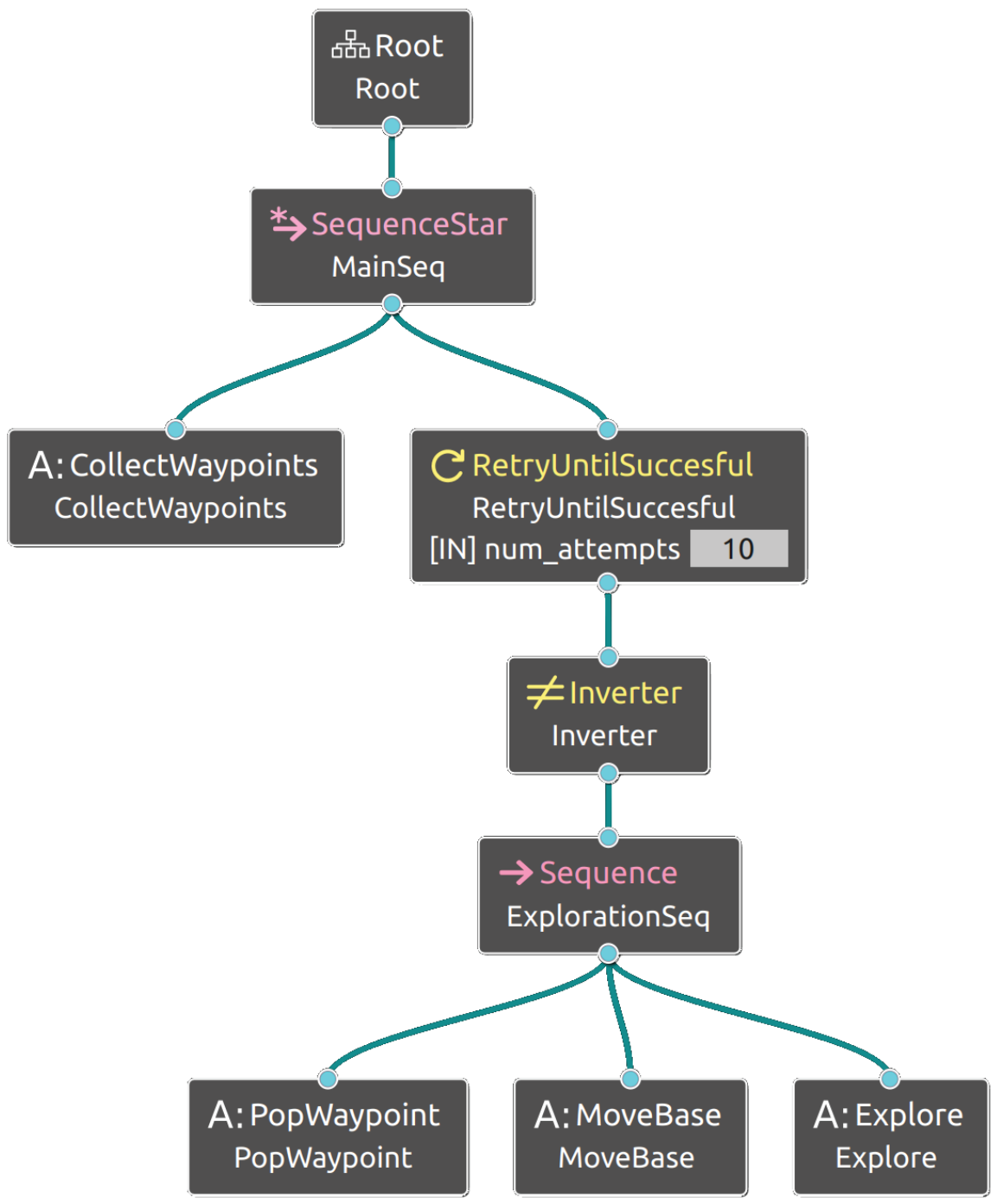

**Figure 1.** An example behavior tree of a health and safety robot inspector from a GitHub project `kmi-robots/hans-ros-supervisor` shown in the Groot editing and animation tool from `BehaviorTree.CPP`

A number of libraries has been developed to implement behavior trees, including common libraries such as `BehaviorTree.CPP` and `py_trees`. In this paper, we explore the concepts offered by behavior tree languages and how these are exploited by the users of these libraries, based on open source projects.

By many researchers, behavior tree languages are praised for their modularity, flexibility, reusability, and ability to express reactive behavior [3, 11, 13–15, 33, 37]. However, none of these claims has been studied upon behavior tree models in real-world projects—the main motivation behind our study.

**Illustrative Example.** Figure 1 presents an example of a behavior tree model of a health and safety inspector robot from the Knowledge Media Institute.[7] The robot performs an exploration sequence of an area. The main operation is placed in the bottom, in the sub-tree under `ExplorationSeq`: it consists of obtaining the next waypoint, moving the mobile base to the waypoint, and exploring the area. If obtaining a new waypoint fails (empty stack) the first task fails, which is inverted into a success by an (`Inverter`) and this means that the sequence of motions have been completed. Otherwise, we keep repeating the same operation (next point, move, explore) up to 10 times, as long as the stack is not empty. The entire computation is placed in an infinite loop of alternating obtaining new waypoints and performing the exploration sequence (`MainSeq`) until the success of all children.

[7] http://kmi.open.ac.uk/

**Behavior Tree Concepts.** In general, a behavior tree is a directed tree with a dedicated root node, with non-leaf nodes called *control-flow nodes* and with leaf nodes called *execution nodes*. A behavior tree is executed by sending signals called *ticks* from the root node down traversing the tree according to the specific semantics of the control-flow nodes. Ticks are issued with a specific frequency [14, 30]. Upon a tick, a node executes a task, which can be a control-flow task or, if a leaf node is ticked, some specific robotic task. The latter classify into actions (e.g., `MoveBase` in Fig. 1) and conditions, which can test propositions (e.g., whether the robot is at its base) used to control task execution. A ticked node returns its status to its parent: (1) success when a task is completed successfully, (2) failure when a task execution failed, and (3) running when a task is still under execution.

The benefit of using behavior trees lies in their ability to express task coordination behavior using a small, but extensible set of control-flow nodes. Most behavior tree languages offer the types sequence, selector, decorator, and parallel, which we will discuss in detail in the remainder (Sect. 4). Our example in Fig. 1 illustrates two sequence nodes (`MainSeq` and `ExplorationSeq`) and two decorator nodes (`Inverter` and `RetryUntilSuccesful`). Intuitively, sequence nodes tick all its children and require all to succeed for the sequence to succeed, while selector nodes only require one to succeed. Decorator nodes allow more complex control flow, including for or while loops. They are also extensible; developers can implement custom decorator nodes. Finally, parallel nodes are generalizations of sequence and selector nodes, allowing custom policies, such as cardinalities specifying the minimum or maximum number of nodes that need to succeed.

# 3 Methodology

We now describe our methodology for identifying and analyzing behavior tree languages (RQ1) and for identifying and analyzing real-world robotic applications using these languages (RQ2 and RQ3).

## 3.1 Behavior Tree Languages

We identified behavior tree languages by searching GitHub for popular behavior tree libraries in `Python` and `C++`, the most used programming languages in robotics. To ensure the libraries' relevance for real-world robotics applications, we focused on maintained libraries that support ROS and applied the following *exclusion criteria*: (1) lack of documentation, (2) out-dated libraries not maintained anymore (last commit older than 2019), and (3) no ROS support.

To understand the modeling concepts offered in behavior trees (RQ1), we studied their relation to concepts found in UML behavior diagrams [27]. Specifically, we systematically compared behavior trees with state machines and activity diagrams. We chose the latter two languages, since they are among the most popular, well-understood, and standardized (via the UML) languages for describing the behaviors. From a robotics user's perspective, behavior trees are becoming an alternative to state machines [8, 13]. Thus, it is natural to compare them. Many other behavior modeling languages are derivatives of state machines or activity diagrams.

For our comparison, we collected behavior tree concepts by means of a thorough literature [9, 14, 35] and library analysis [21, 39], then carefully mapped (based on their semantics) these concepts to related concepts found in the other two UML languages. In this process, we focused on behavior tree concepts and whether state machines and activity diagrams offer direct support for similar concepts or whether they need to be expressed indirectly. Our analysis was iterative, to ensure a proper reflection of the concepts in the different languages.

## 3.2 Behavior Tree Models

For the identified behavior tree libraries, we investigated how they are used in the source code of robotics projects. In `BehaviorTree.CPP`, the term *main_tree_to_execute* refers to the entry point tree in the XML source code, while the term *py_trees_ros* is used to import the language `py_trees_ros`. Both terms must be used in the source code of targeted languages. To this end, we created a `Python` script to mine GitHub repositories using those terms for a simple text-match in source code and `GitHub`'s code search API.[8] After mining GitHub for open-source projects, we manually explored the found projects to identify the relevant ones. To focus on behavior tree models used in real robotic projects, we excluded projects belonging to (1) a tutorial or to (2) a course.

To understand the use and characteristics of behavior tree models (RQ2 and RQ3), we analyzed the identified projects. We explored manually and semi-automatically; the latter by calculating metrics to understand how behavior tree concepts are used in the selected projects from a statistical perspective. Those metrics are:

- The size of the behavior tree (BT.size): number of all nodes excluding the root node.
- The tree depth (BT.depth): number of edges from the root node to the deepest node of the tree [17].
- Node type percentage (N.pct):the frequency of a node type with respect to the total number of nodes.
- Average branching factor (ABF): the average number of children at each node.

To calculate BT.size and N.pct, we extracted a function name for each node type based on the libraries' documentation, then used a `Python` script to count the number of text matches. For leaf nodes, automatic counting was only possible for libraries imposing a specific structure on the leaf nodes; otherwise, we counted manually. We manually calculated BT.depth and ABF, since we needed to manually extract the models anyway. Note that, while these metrics capture core structural aspects of the models, answering RQ2 and RQ3, we specifically focus on reuse as one of the major issues in robotics software engineering [25, 26].

[8]https://github.com/PyGithub/PyGithub

**Table 1.** Behavior tree languages identified (we analyzed the implementations of the first three, which are in bold)

| Name | Language | ROS | Doc. | Last commit |
|---|---|---|---|---|
| **`BehaviorTree.CPP`**<br>github.com/BehaviorTree/BehaviorTree.CPP | C++ | yes | [21] | 2020/05/16 |
| **`py_trees`**<br>github.com/splintered-reality/py_trees | Python | no | [39] | 2020/03/10 |
| **`py_trees_ros`**<br>github.com/splintered-reality/py_trees_ros | Python | yes | [40] | 2020/02/25 |
| BT++<br>github.com/miccol/ROS-Behavior-Tree | C++ | yes | [10] | 2018/10/22 |
| Beetree<br>github.com/futureneer/beetree | Python | yes | N/A | 2016/03/14 |
| UE4 Behavior Tree<br>docs.unrealengine.com/en-US/Engine/ArtificialIntelligence/BehaviorTrees | UnrealScript | no | [22] | N/A |

We manually inspected the models. Looking at each model, we identified different usage patterns depending on the used language. We were able to use a visual editor shipped with one of the identified libraries (Groot for `BehaviorTree.CPP`, explained shortly) where the behavior tree language is re–alized as an external DSL. The other identified library (`py_trees_ros`, explained shortly) constituted an internal DSL, where we needed to manually extract the model from the source code, identifying the respective library API calls constructing the model. There, we considered every tree with a root node as a behavior tree model.

# 4 Behavior Tree Languages (RQ1)

Table 1 lists the implementations of behavior tree languages identified and considered in this study; five from the robotics community and one from outside. This section focuses on analyzing the implementations in the first three rows, set in bold font. Among the languages relevant for robotics, these three were actively developed when we checked (2020/05/16). Together they support ROS systems implemented in Python and C++, the two most popular programming languages in the robotics community. The `py_trees` library, the main behavior tree implementation in the Python community, does not directly target ROS, but robotics in general. A popular extension, `py_trees_ros`, provides bindings for ROS. Since `py_trees` and `py_trees_ros` are similar, with the only difference of ROS packaging, we decided to include `py_trees` in the language analysis even though it does not support ROS directly.

We decided to discard the remaining three languages from our analysis. BT++ is now obsolete, superseded by `Behavior-Tree.CPP` after the developer of BT++ joined the latter as a contributor. Beetree is an inactive experiment, now abandoned. *Unreal Engine 4 (UE4) Behavior Tree*, probably the world's most used behavior tree dialect, is a well-documented library with a graphical editor to model intelligent actor behavior in games. However, the game development use case impacts the implementation. It emphasizes event-driven programming rather than time-triggered control, which is the major concern in robotics. Since we focus on robotics and not computer games, we will not discuss it any further.

## 4.1 Language Subject Matter

Behavior trees can be seen as graphical models that are shaped as trees, representing tasks for execution by an agent. Robotics and gaming are domains where autonomous agents are frequently programmed [14]. A model consists of composite control flow nodes that coordinate how the basic action nodes should be scheduled by the agent. The visual presentation of the main node types is summarized in Fig. 2 as used in robotics and games [14, 30, 35]. The four basic categories of control flow are: Sequence, Selector, Parallel, and Decorator. The two basic execution nodes are Action and Condition. Each tree has a designated Root node. To illustrate the abstract syntax, we also provide a meta-model we reverse-engineered from `BehaviorTree.CPP`'s XML format in Fig. 3 and most of these concepts are explained in detail in Table 2.

Table 2 summarizes the key aspects of our analysis of the concepts and benefits of behavior trees and their comparison with UML state machines and activity diagrams. The left-most column names concepts pertinent to the behavior tree languages, either due to inclusion or a striking exclusion from behavior tree languages. The last two columns comment briefly on how the respective concept is handled in the UML languages.

## 4.2 Language Design and Architecture

Turning our attention to how behavior tree languages are implemented from the language design perspective, the first striking observation is that both languages are predominantly distributed as libraries, not as language tool chains, or modeling environments. `BehaviorTree.CPP` is implemented as a `C++` library, packaged as a ROS component, easy to integrate with a ROS-based codebase [20]. In contrast, `py_trees` is a pure `Python` library. It has an extension `py_trees_ros` which packages `py_trees` as a ROS package and adds ROS-specific nodes.

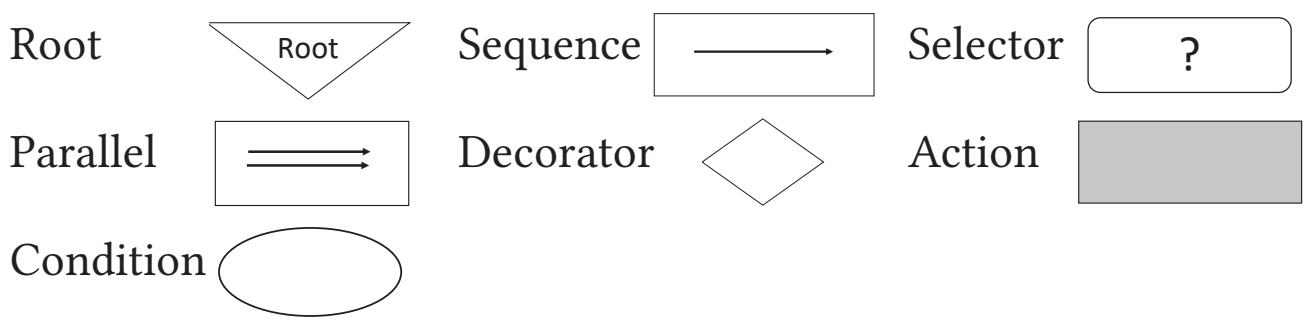

**Figure 2.** Behavior Trees node types (visual syntax)

**Table 2.** Selected key language concepts in behavior trees, and a comparison with UML diagrams

| Concept/aspect | Behavior trees | Activity diagrams | State diagrams |
|---|---|---|---|
| programming model | Synchronized, time-triggered, activity-based. Reactive programming can be implemented to an extent using tick and re-ordering sub-trees. | Asynchronous, reactive, explicit control-flow | Synchronous, reactive, explicit control-flow |
| **simple nodes** | Execute **actions** (arbitrary commands, both instantaneous and long-lasting) or evaluate **conditions** (value translated to success/failure). | Basic activity | Basic action |
| exit status | Each node reports success, failure, or an in-operation state ("running") each time it is triggered. Status report causes the computation (the traversal) to advance to the next node. | Completion of an activity advances state like in BTs, failures modeled by exceptions/handlers | No direct support, control-flow mostly driven by message passing, not by activity flow |
| **composite nodes** | Define hierarchical traversal, the control-flow for each epoch (tick). Sequentially composed. Nodes may start concurrent code though. | Nested activities | Nested states, both sequential and parallel |
| root | Serves as entry point for every traversal. Has exactly one child node. The root node is re-entered at every epoch. | Initial node, possibly a root activity, initialized only once | Initial state, possibly a root state, initialized only once |
| sequence | Trigger children in a sequence until the first failure. If no failure return success, otherwise fail. | No direct support, detect failure with exception handler | No direct support, use a parent transition on failure |
| selector | Trigger children in a sequence until the first success. If no success return failure, otherwise succeed. | No direct support, difficult to model with exceptions | No direct support, use a parent transition on success |
| parallel | Generalize sequence/selector with a policy parameter. Several polices available, e.g. meeting a minimum number succeeding children. | No direct support | No direct support |
| goto (jumps) | No general jump construct, the computation always traverses the tree. | Supported | Supported |
| **decorators** | | | |
| inverter | Invert the Success/Failure status of the child | No expl. inversion operator | No expl. inversion operator |
| succeed | Return success regardless of status returned by the child | No expl. support for status | No expl. support for status |
| repeat | Trigger the child node a set number of times, then succeed. Fail if the child fails. | No direct support. Encode with loops + counter/guard | No direct support. Encode with loops + counter/guard |
| retry | Run the child node and retry it immediately if it fails for a maximum number of times, otherwise succeed. | No direct support. Encode with loops + counter/guard | No direct support. Encode with loops + counter/guard |
| dynamicity | Runtime modifications of model (re-ordering of nodes) possible due to the dynamic nature of the implementation | The syntax (order of execution) is fixed at run-time. | The syntax (order of execution) is fixed at run-time. |
| openness | New nodes and operators implemented by users as needed | Not possible (closed) | Not possible (closed) |

**Concrete Syntax.** The syntactic elements of behavior trees are presented graphically in Fig. 2. Fig. 1 showed an example model in a slightly different concrete syntax. Both dialects come with ways to visualize models as graphs, and `BehaviorTree.CPP` even has a graphical editor and a visual runtime monitor for its models called Groot (which which the graphical representation of a behavior tree was visualized in Fig. 1).

Nevertheless, it is important to understand that behavior trees are *not* a visual modeling language in a traditional sense. First, in both libraries, the models are constructed in a text editor, in a mixture of C++, respectively Python. Second, the models are constructed directly in abstract syntax, by instantiating and wiring abstract syntax types. For convenience, and to support Groot, `BehaviorTree.CPP` offers an XML format, which can be used to write the tree syntax in static files. This file is interpreted at runtime, and an abstract syntax tree is constructed from it dynamically. Third, crucially, the types of nodes (and, thus, the XML file in `BehaviorTree.CPP`) do not constitute the entire meaning of the model. An important part of the model is embedded in C++/Python code that is placed in the methods of the custom node classes. This part of the model is neither modifiable nor presentable in the graphical tools.

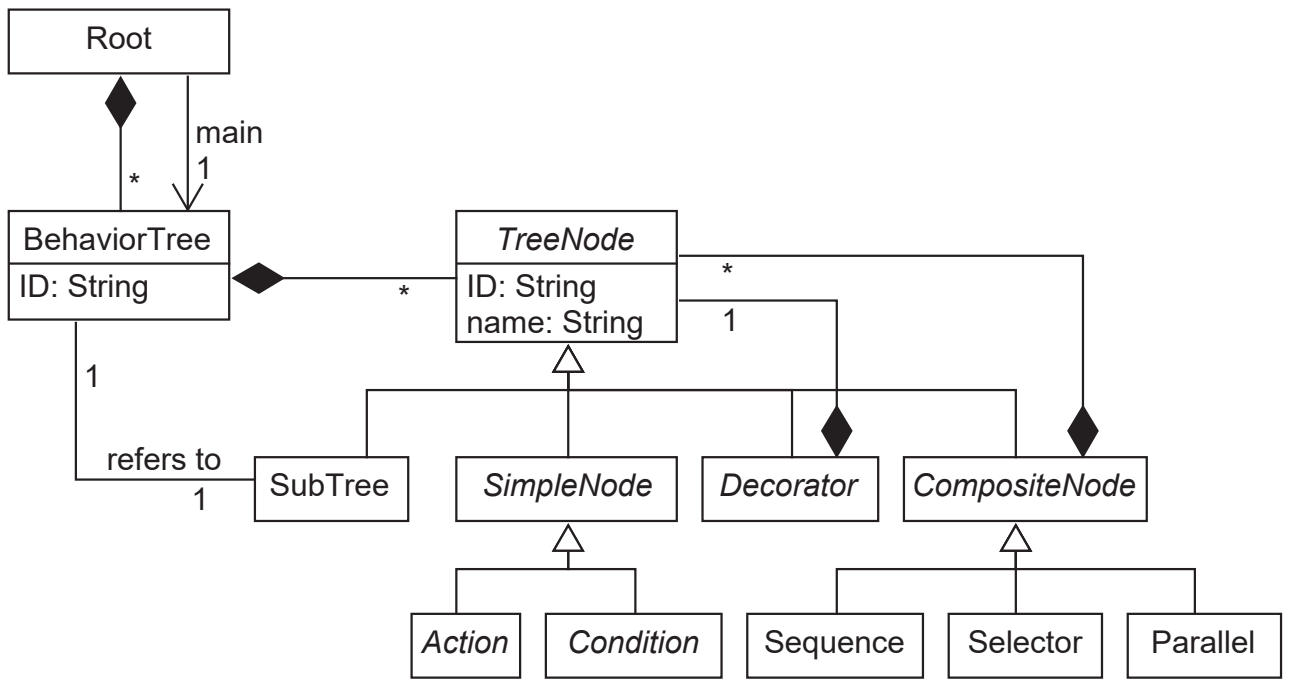

**Figure 3.** A meta-model for `BehaviorTree.CPP` (reverse-engineered from its XML format)

Finally, recall that `BehaviorTree.CPP` is realized as an external DSL through Groot and the XML-like format, while `py_trees_ros` constitutes an internal DSL, since it does not have similar tools. From our experience analyzing their models (cf. Sect. 5), we can confirm that the `BehaviorTree.CPP` models are much easier to comprehend, and the availability of its visual editor Groot has made it faster to analyze the behavior tree models than `py_trees_ros` models.

**Semantics of Behavior Trees.** The variant of behavior trees used in robotics is predominantly a *timed-triggered activity-based* behavioral modeling language. The computation consists of activities that have duration, like in activity diagrams. Unlike in activity diagrams, the main control loop does not shift control tokens or states around. Instead, it triggers the entire model at (typically) fixed intervals of time like a circuit. Every tick (or epoch) triggers a traversal of the entire tree, with diversions introduced by various types of nodes. The traversal can start new activities, evaluate conditions, access state, and execute basic actions for side effects. Reactive programming seems not to be supported first-class, despite reappearing statements to the contrary,[9] but can be simulated by sufficiently high-frequency model execution.

The model has a global storage called *blackboard*, which is a key-value store. No scopes are supported; all keys are global. The blackboard is used for communicating, both within the model and with the rest of the system. The model and the system read and update the blackboard asynchronously.

**Simple Nodes.** Simple nodes, or leaves in the syntax tree, are either conditions or actions. Actions realize the basic computation in the model. Users of the language need to implement custom action nodes—classes obeying the Action interface that contain Python or C++ code to be executed whenever a node is ticked. Conditions calculate a value of a Boolean predicate and convert it to a success or failure value.

Simple nodes, and by propagation also composite nodes, return an explicit exit status, which can be a success, a failure, or information that the action is still running. These values propagate upwards during the tree traversal according to the semantics of composite nodes, discussed below. The semantics resembles that of a logical circuit, a neural network, a flow diagram, or a computation in the failure monad known in pure functional programming (but the modeling language is obviously far from pure). The model receives results from simple nodes and converts them through the network.

The simplest Action nodes are synchronous, so they terminate quickly and return success or failure immediately. Asynchronous nodes may also return a 'running' status and use some form of concurrency to continue operation. The execution engine will attempt to trigger them at the next epoch again. The design of behavior tree languages does not prescribe the model of concurrency, and implementations vary. For instance, `BehaviorTree.CPP` implements asynchronous nodes using coroutines [16]. A node that is not ready to terminate can yield to the engine, and be restarted again at the next epoch. This directly gives interleaving concurrency, but can give true concurrency if the executed code uses threads or parallel processes (which it would typically do in ROS). Coroutine semantics is extremely rare in modeling languages. It was present in Simula [18]. Statecharts had a weak form of coroutines as 'history states' [28], and more recently they were used cooperatively in ABS [32]. It is interesting that this semantics is coming back, thanks to programming languages re-discovering it. It is now supported in Python and included in the 2020 C++ specification.

> Observation 1. Implementations of behavior tree languages support both interleaving and true concurrency using threads and coroutines. The model of concurrency is not defined strictly in the language, but instead, left largely to the users.

**Composite Nodes.** Composite nodes are internal nodes of a behavior tree. Their main function is to define the order of traversal at every time epoch (at every trigger). Unlike for simple nodes, which need to be implemented by the user, the language provides a range of predefined composite nodes. The *root* node is the composite node that serves as an entry point for every traversal, it contains another node as the body. This node is re-entered to start every traversal. UML languages do not have an explicit notion of ticks and of reoccurring traversals. Both Activity Diagrams and State Diagrams have initial nodes and a possibility to nest the model in a root node, but their initial nodes are only started once at the beginning of the model execution, revisited only if the control-flow gets there.

[9] For example, the `py_trees` documentation states that the language provides *a good blend of purposeful planning towards goals with enough reactivity to shift in the presence of important events*; https://py-trees.readthedocs.io/en/devel/background.html

**Table 3.** Behavior tree concepts and corresponding language elements in BehaviorTree.CPP and py_trees

| Concept | BehaviorTree.CPP | py_trees |
|---|---|---|
| Simple Node | subclasses of ActionNode ConditionNode | behaviour.Behaviour |
| Composite | subclasses of ControlNode | classes in composites |
| Sequence | Sequence, SequenceStar ReactiveSequence | composites.Sequence |
| Selector | Fallback, FallbackStar ReactiveFallback | composites.Selector composites.Chooser |
| Decorator | subclasses of DecoratorNode | classes in decorators |
| Parallel | ParallelNode | composites.Parallel |

A *sequence* node triggers (visits) all children until the first failure (similar to a *forall* higher order function, which is standard in many programming languages). A *selector* node triggers all children until the first success (similar to *exist*). A *parallel* node is really a misnomer. It does not execute nodes concurrently, but generalizes sequence and selector to a range of *policies*; that is, the subset of children that shall succeed or fail.

Since the execution is always a traversal of the entire tree, there is no direct support for jumps (goto). Instead, composite nodes can affect the traversal locally, in stark contrast to both activity diagrams and state diagrams. In these languages, a typical change of control allows an arbitrary change of state, often cross-cutting the syntax tree.

**Decorators.** Decorators are unary composite nodes (only one child). They decorate the sub-trees and modify their data or control flow. An *Inverter* flips the return status of a child between success and failure. A *Succeeder* always succeeds regardless the status returned by its child node. A *Repeat* node, which is stateful, acts like a for-loop: it continues to trigger the child for a given number of ticks. It increments an internal counter at every trigger. The node succeeds (and resets the counter) on hitting a set bound. It fails (and resets the counter) if the child fails. A *Retry* node resembles a repeat node. Its main goal is to make a flaky node succeed. Like Repeat it can run a node up to a set number of times, but unlike Repeat, it only retries when a node fails and it retries immediately without waiting for the next epoch. It fails if the child failed in a given number of attempts.

> Observation 2. The conceptual scope and semantics of behavior tree languages differ significantly from the modeling languages in UML. Behavior trees gather a number of constructs based on patterns that, according to users and developers, are frequently found in high-level control of autonomous systems.

The above discussion is based on a broad description of behavior languages extracted from the available literature and documentation of py_trees, py_trees_ros, and BehaviorTree.CPP [21, 39]. Table 3 presents the names of the basic Behavior Trees concepts in the two dialects.

**An interpreter or a compiler?** Both dialects are interpreted. Once the abstract syntax tree is constructed, the user is supposed to call a method to trigger the model once, or to trigger it continuously at a fixed frequency. This does not seem to depart far from other applications of models-at-runtime [4, 5]. BehaviorTree.CPP uses template metaprogramming instead of code generation, which allows to offer a bit of type-safety when implementing custom tree nodes, without exposing users to any specialized code-generation tools. Using the library appears like using a regular C++ library. As expected, no static type safety is offered in py_trees.

**Openness.** The openness and indefiniteness of behavior trees are probably their most interesting aspects, after the time-triggered coroutine-based model of computation. Others have also noticed this in the context of variability in DSLs [41]. Both languages are unusually open. BehaviorTree.CPP is technically an external DSL, but its implementation exposes aspects of dynamic internal DSLs. The programmer can both create models in XML (external, static), and create new node types or modify the shape of the syntax tree at runtime (dynamic). py_trees is an entirely dynamic DSL, where new node types and Python code can be freely mixed, like in internal DSLs.

Unlike in Ecore[10] or UML, the language meta-model is *not fixed*. The basic implementation provides the meta-classes for composite nodes, while it leaves the simple nodes abstract or only gives them bare bones functionality (cf. Fig. 3). A user of the language is expected to first extend the meta-model by implementing the basic action nodes, then link them together in a syntax tree, possibly using an external XML file. This practice vaguely resembles stereotyping [27]. Obviously, a user of Ecore can extend the meta-model classes and give them new functionality at runtime as well, however such use of Ecore is considered advanced and is seen rather rarely. The difference is that of degree: there is essentially no way to consider using Behavior Trees without creating custom nodes.

This design pragmatically supports openness of the language and makes adaptation to diverse scenarios in robotics easy. The openness seems to be required due to a lack of agreement in the robotics community about the ideal control model for robot behavior. Since this question is likely to remain open for a long time, the design allows users to adapt the language as they see fit when building robots.

[10]https://www.eclipse.org/modeling/emf/

**Prerequisites (User Demographics).** The open nature of Behavior Trees means that the experience of building and debugging models resembles very much language-oriented programming as practiced in the modeling and language design research community. One constantly deals with meta-classes, composing them, traversing them, etc. Anybody familiar with building DSLs on top of Ecore or similar frameworks will definitely experience a *déjà vu*, when using either `py_trees` or `BehaviorTree.CPP`.

Given that many robotics engineers, and many ROS users, lack formal training in computer science and software engineering [2], it is surprising to us that this design seems to be well received in the community. Even within software engineering, language implementation and meta-programming skills are often considered advanced. Yet, using Behavior Trees requires such skills. A challenge for the modeling community is lurking here: to design a Behavior Trees language that, while remaining flexible and easy to integrate with large and complex existing code bases, is much easier to use for a regular robotics programmer.

> Observation 3. The flexibility and extensibility of Behavior Trees require language-oriented programming skills from robotics developers. The software-language engineering community could contribute by designing an accessible, but still flexible, dialect of Behavior Trees.

**Separation of Concerns.** Behavior Trees are platform-specific models (PSMs) built as part of a specific robotics system to control behaviors at runtime. The models are used to simplify and conceptualize the description of behavior. The ability to reuse the same models with other hardware or similar systems is not (yet!) a primary concern. Behavior Trees not only are PSMs, but tend to be very tightly integrated with the system. Custom nodes tend to refer to system elements directly and interact with the system API. As a result, it is hard to use these models separately from the robot. While Groot can visualize a standalone XML file of a model, a working build environment of ROS is needed just to visualize the syntax of a `py_trees_ros` model. This may mean not only an installation of suitable Python and ROS libraries, but, for example, a working simulation of the robot, or even the hardware environments. You need to launch the system and inject a visualization call to inspect the model!

It is in principle possible with both libraries to build models that are completely decoupled from the system. It suffices to route all communication with the system via the blackboard. `BehaviorTree.CPP` provides dedicated XML primitives for this purpose, allowing the entire behavior to be programmed in XML, provided the rest of the system can read from and write to the blackboard. This separation allows models to be processed outside the system for visualization, testing, grafting into other systems, and so on. We definitely think this is a good architectural practice to follow. Nevertheless, it is not what we observed in real-world models (cf. Sect. 5). Most models mix the specification of behavior deeply with its implementation, making separation virtually impossible.

> Observation 4. Behavior tree models tend to be deeply intertwined with behavioral glue code linking them to the underlying software system. This makes operating on models outside the system difficult, hampering visualization, testing, and reuse.

# 5 Behavior Tree Models (RQ2 & RQ3)

We identified 75 behavior tree models belonging to 25 robotic projects, as summarized in Table 4. Their domains are:

- navigation and verbal communication (gizmo, neuronbot2_multibot, vizzy_playground, vizzy_behavior_trees, MiRON-project, behavior_tree_roscpp, BT_ros2);
- pick-and-place (stardust, refills_second_review, pickplace, mobile_robot_project, mecatro-P17);
- serving robot (Pilot-URJC, robocup2020, BTCompiler, Yarp-Smart-Soft-Integration, carve-scenarios-config);
- real-time strategy and inspection (roborts_project, robotics-player, Robotics-Behaviour-Planning);
- health and nursing home (hans-ros-supervisor, bundles);
- testing submarine hardware (Smarc_missions, sam_march);
- drone-based parcel delivery (dyno).

**RQ2. Use of Behavior Tree Language Concepts.** We measured the metrics explained in Sect. 3.2 on `py_trees_ros` and `BehaviorTree.CPP` projects. Table 4 presents these metrics under model characteristics. In general, we noticed a large variation in BT.size among models (11% of models have a BT.size > 50, 56% ≥ 10, and 33% of models have BT.size < 10).

In addition, 66% of total node types were leaf nodes (1, 228 out of 1, 850 total node types), while composite nodes acquired 34% of total node types. Since leaf nodes are dominated in the model, we decided to explore the usage of composite concepts against each other to have a better understanding of how the concepts are used. Table 5 summarizes the usage of composite nodes for each studied project models (as of 2020/07/16).

Most of the composite nodes in our projects are of type Sequence (53% with `py_trees_ros`, 57% with `BehaviorTree.CPP`) and Selector (28% and 19% respectively). The Parallel concept was not used much, only 7% of total composite nodes. (The reader might recall that it is not any more concurrent than Sequence and Selector.) This perhaps explains why standard libraries of programming languages normally do not include generalizations of existential and universal quantifier functions (exists and forall)—these use cases seem to be rare. The re-entrant nature of the behavior tree language allows to use Parallel to wait until a minimum number of sub-trees succeed. This however does not seem to be used as often as we expected.

**Table 4.** Subject projects identified from GitHub that use behavior tree models to define robot behavior. The average of BT.size and BT.depth were taken for projects with multiple models.

| Project Name & GitHub Repository | Last Commit | Lang. | Models | Model Description | Model Characteristics BT.size | BT.depth | N.pct | ABF |
|---|---|---|---|---|---|---|---|---|
| sam_march<br>KKalem/sam_march | 2019/03/08 | py | 1 | Model re-use sub-trees and some functions in different branches. Exploits modularity by reusing sub-trees | 53 | 9 | Comp: 38%<br>Leaf: 62% | 2.5 |
| mobile_robot_project<br>simutisernestas/mobile_robot_project | 2019/10/07 | py | 1 | Model reuses same sub-tree in different places in the tree. Shows modularity of BTs. | 30 | 5 | Comp: 40%<br>Leaf: 60% | 2.5 |
| smarc_missions<br>smarc-project/smarc_missions | 2020/02/17 | py | 2 | Exploits modularity by reusing sub-trees. puts ROS data into blackboard shared among sub-trees. | 29 | 7 | Comp: 36%<br>Leaf: 64% | 2.7 |
| dyno<br>samiamlabs/dyno | 2018/10/11 | py | 2 | Two models with same structure (e.g., composite nodes, depth, size), differ in leaf nodes and passed parameters. Illustrates reuse among BTs. | 29 | 6 | Comp: 38%<br>Leaf: 62% | 3 |
| gizmo<br>peterheim1/gizmo | 2019/02/22 | py | 8 | Multiple models with same structure (e.g., tree depth, model size) but minor changes for nodes type. Shows use of different node types to implement same mission with similar BT structure. | 17 | 5 | Comp: 38%<br>Leaf: 62% | 2.4 |
| roborts_project<br>Taospirit/roborts_project | 2019/11/26 | py | 1 | Complex robot tasks, but simple model. Multiple sequence sub-trees ends with leaf nodes executing functions with the actual complex tasks. | 16 | 3 | Comp: 38%<br>Leaf: 63% | 2.5 |
| robotics-player<br>braineniac/robotics-player | 2018/06/29 | py | 1 | Each tasks is modeled using sequence sub-trees. Simple model, actions are executed in leaf nodes by reading and writing to blackboard. | 12 | 4 | Comp: 33%<br>Leaf: 67% | 2.8 |
| Robotics-Behaviour-Planning<br>jotix16/Robotics-Behaviour-Planning | 2019/10/05 | py | 3 | Models with similar tree structure, but different parameters are passed depending on the task. | 12 | 3 | Comp: 23%<br>Leaf: 77% | 4.4 |
| refills_second_review<br>refills-project/refills_second_review | 2020/02/19 | py | 1 | Simple model. A sub-tree executes sequence of tasks (implemented via functions), which use blackboard to read sensors values (e.g., shelf ID). | 8 | 4 | Comp: 38%<br>Leaf: 63% | 2.3 |
| pickplace<br>ipa-rar/pickplace | 2020/04/01 | C++ | 1 | Model reuses sub-tree concept in different ways; same sub-tree in different places and sub-trees almost identical but have minor difference (setBlackboard node added). Shows modularity and reusability in BTs. | 85 | 8 | Comp: 32%<br>Leaf: 68% | 3.0 |
| stardust<br>julienbayle/stardust | 2019/06/09 | C++ | 4 | Among our largest models found. Exploits modularity and reusability of sub-trees. | 56 | 8 | Comp: 39%<br>Leaf: 61% | 2.7 |
| neuronbot2_multibot<br>skylerpan/neuronbot2_multibot | 2020/07/12 | C++ | 2 | Both models are similar but passing different parameters to the tree. Exploits modularity of sub-trees. | 54 | 10 | Comp: 50%<br>Leaf: 50% | 2.1 |
| mecatro-P17<br>alexandrethm/mecatro-P17 | 2019/06/01 | C++ | 11 | Models with multiple sub-trees for different sequential tasks. Some sub-trees reuse structure, but other parameters passed based on task. | 49 | 4 | Comp: 20%<br>Leaf: 80% | 4.7 |
| Yarp-SmartSoft-Integration<br>CARVE-ROBMOSYS/Yarp-SmartSoft-Integration | 2019/04/23 | C++ | 1 | Shows combination of Selector and condition nodes to keep track of robot surrounding environment before executing action. | 31 | 7 | Comp: 42%<br>Leaf: 58% | 2.2 |
| bundles<br>MiRON-project/bundles | 2020/07/10 | C++ | 5 | The different models take advantage of behavior trees modularity using the same sub-trees. | 16 | 5 | Comp: 43%<br>Leaf: 57% | 1.8 |
| BTCompiler<br>CARVE-ROBMOSYS/BTCompiler | 2019/04/09 | C++ | 8 | Shows combination of Selector and condition nodes to keep track of a robot surrounding environment before executing action. Two models similar to Yarp-SmartSoft-Integration, illustrating cross-project reuse. | 15 | 6 | Comp: 40%<br>Leaf: 60% | 2.1 |
| BT_ros2<br>Adlink-ROS/BT_ros2 | 2020/06/12 | C++ | 2 | The two models exploit modularity and reusability of sub-trees. | 14 | 7 | Comp: 54%<br>Leaf: 46% | 1.6 |
| vizzy_behavior_trees<br>vislab-tecnico-lisboa/vizzy_behavior_trees | 2020/04/02 | C++ | 7 | The models show different ways to implement the same activity with or without sub-trees. | 13 | 6 | Comp: 45%<br>Leaf: 55% | 1.8 |
| hans-ros-supervisor<br>kmi-robots/hans-ros-supervisor | 2018/10/11 | C++ | 1 | Simple model with only two sub trees. | 8 | 5 | Comp: 50%<br>Leaf: 50% | 1.6 |
| Pilot-URJC<br>MROS-RobMoSys-ITP/Pilot-URJC | 2020/06/10 | C++ | 2 | Models contain customized nodes implementation (developers implemented a customized node type) for sequence and decorator nodes. | 8 | 4 | Comp: 40%<br>Leaf: 60% | 2.6 |
| vizzy_playground<br>vislab-tecnico-lisboa/vizzy_playground | 2020/04/05 | C++ | 6 | Shows simple behavior tree models using combinations of sequence and action nodes. | 8 | 3 | Comp: 32%<br>Leaf: 68% | 2.0 |
| behavior_tree_rosC++<br>ParthasarathyBana/behavior_tree_rosC++ | 2020/06/24 | C++ | 1 | Simple behavior tree model showing the usage of combining sequence and condition nodes. | 7 | 3 | Comp: 43%<br>Leaf: 57% | 1.8 |
| MiRON-project<br>ajbandera/MiRON-project | 2020/02/21 | C++ | 1 | Simple behavior tree model. | 7 | 2 | Comp: 14%<br>Leaf: 86% | 6 |
| robocup2020<br>IntelligentRoboticsLabs/robocup2020 | 2020/03/25 | C++ | 2 | Simple behavior tree model with customized nodes implementation. | 7 | 2 | Comp: 23%<br>Leaf: 77% | 3 |

**Table 5.** Usage of different composite nodes to the total of composite nodes in the identified robotic projects and in total for all projects

| Project Name | | Sequence | Selector | Decorator | Parallel |
|---|---|---|---|---|---|
| | | **Composite Nodes** | | | |
| roborts_project | **py_trees_ros** | 83% | 17% | 0% | 0% |
| refills_second_review | | 67% | 0% | 33% | 0% |
| gizmo | | 58% | 17% | 6% | 19% |
| smarc_missions | | 57% | 29% | 0% | 14% |
| sam_march | | 55% | 25% | 0% | 20% |
| robotics-player | | 50% | 50% | 0% | 0% |
| dyno | | 45% | 27% | 18% | 9% |
| Robotics-Behaviour-Planning | | 38% | 63% | 0% | 0% |
| mobile_robot_project | | 25% | 67% | 8% | 0% |
| robocup2020 | **BehaviorTree.CPP** | 100% | 0% | 0% | 0% |
| MiRON-project | | 100% | 0% | 0% | 0% |
| pickplace | | 81% | 11% | 0% | 7% |
| BT_ros2 | | 80% | 7% | 13% | 0% |
| vizzy_playground | | 73% | 20% | 7% | 0% |
| mecatro-P17 | | 70% | 0% | 19% | 11% |
| behavior_tree_roscpp | | 67% | 33% | 0% | 0% |
| neuronbot2_multibot | | 59% | 11% | 26% | 4% |
| vizzy_behavior_trees | | 53% | 23% | 13% | 13% |
| Pilot-URJC | | 50% | 17% | 33% | 0% |
| hans-ros-supervisor | | 50% | 0% | 50% | 0% |
| BTCompiler | | 46% | 54% | 0% | 0% |
| stardust | | 44% | 16% | 38% | 2% |
| carve-scenarios-config | | 38% | 62% | 0% | 0% |
| Yarp-SmartSoft-Integration | | 38% | 63% | 0% | 0% |
| bundles | | 37% | 20% | 31% | 11% |
| Share in population of all models | | 56% | 21% | 16% | 7% |

Decorators are used relatively rarely in `py_trees_ros` models, they constitute 6% of the composite nodes. This is likely explained by the fact that it is easier to apply the transforming operations directly in the Python code, using Python syntax, than elevating it to behavior tree abstract syntax constructors. The situation is different with `BehaviorTree.CPP`, where decorators are used almost three times as often (19% of composite nodes). Here, the benefit of using the decorators (data-flow operators) of behavior tree instead of C++ allows them to be visualized and monitored in the graphical editor (Groot). No such tool is available for `py_trees`, so likely bigger parts of the model may "leak" to the code. This demonstrate that Behavior trees users often have a choice of what is in scope and what out of scope for a model. This is a property that clearly distinguishes GPLs from DSLs. Yet, in our experience, the skill of deciding the model scope and the precision level is rarely discussed in teaching and research literature.

Finally, we have observed that none of the models implement their own custom nodes. They relay on the extensibility of behavior trees using new custom operator (decorators). By using the available off-shelf decorators in `BehaviorTree.CPP` and `py_trees_ros`, they were sufficient to create a custom behavior to change an action/condition status, or customize an action length, e.g. want to execute an action without waiting, retry an action $n$ times before given up, or repeat an action $n$ times.

Going back to Fig. 1, the decorator `RetryUntilSuccesful` was used to create a conditional loop that executes the subtree under (`ExplorationSeq`) 10 times, unless the task fails, which is inverted into a success by an (`Inverter`). The developers were able to model this without having to use while-loop or a similar general control-flow structure in the script.

> Observation 5. The studied Behavior tree languages offer a range of concepts that are well suited to roboticists, but the offered concepts usage might differ according to the language.

**RQ3. Characteristics of behavior tree models.** We already presented core structural characteristics of our models in Table 4. We now focus on reuse as one of the major issues in robotics software engineering [25, 26]. In fact, our qualitative analysis of the models shows that reusing parts of the trees plays a major role.

Reusing refers to the ability to divide a mission into subtasks represented by sub-trees or functions and reusing them in the same models or across models. The creators of our models tend to divide them into sub-tasks, which are represented by sub-trees or actions that can be easily separated and recombined like building blocks. They can be re-used in other models when they share similar activities, improving the efficiency of modeling.

We observed three patterns of reusing in the studied behavior tree models: *reuse by reference*, *reuse by clone-and-own* [19], and *reuse by reference through file inclusion.*

59% of behavior tree models exploit *reuse by reference* in their models, and in the projects with multiple models, developers even reuse across the different models (33% of projects). Developers implemented reuse by reference mostly by creating a sub-tree for a repeated activity, then re-using it by reference in multiple branches in the model after passing the new values for its parameters (usually writing a new value to a blackboard). Another implementation is by defining a leaf node as a function in an external file (header files), then reusing it by reference after passing new values to its parameters. Figure 4 shows an excerpt from one of our models, presenting the different tasks for a robot in a retirement home. The red box highlights an example of reuse by reference, where the developer wrapped the moving activity in the sub-tree (`Recharge`) and reused it in multiple parts of the model. Another example of reuse by reference, but for a leaf node, is shown in the action `moveRoboterPosition`, where it was used in multiple parts in the model, only changing the parameters' values (name, approachRadius, x, and y).

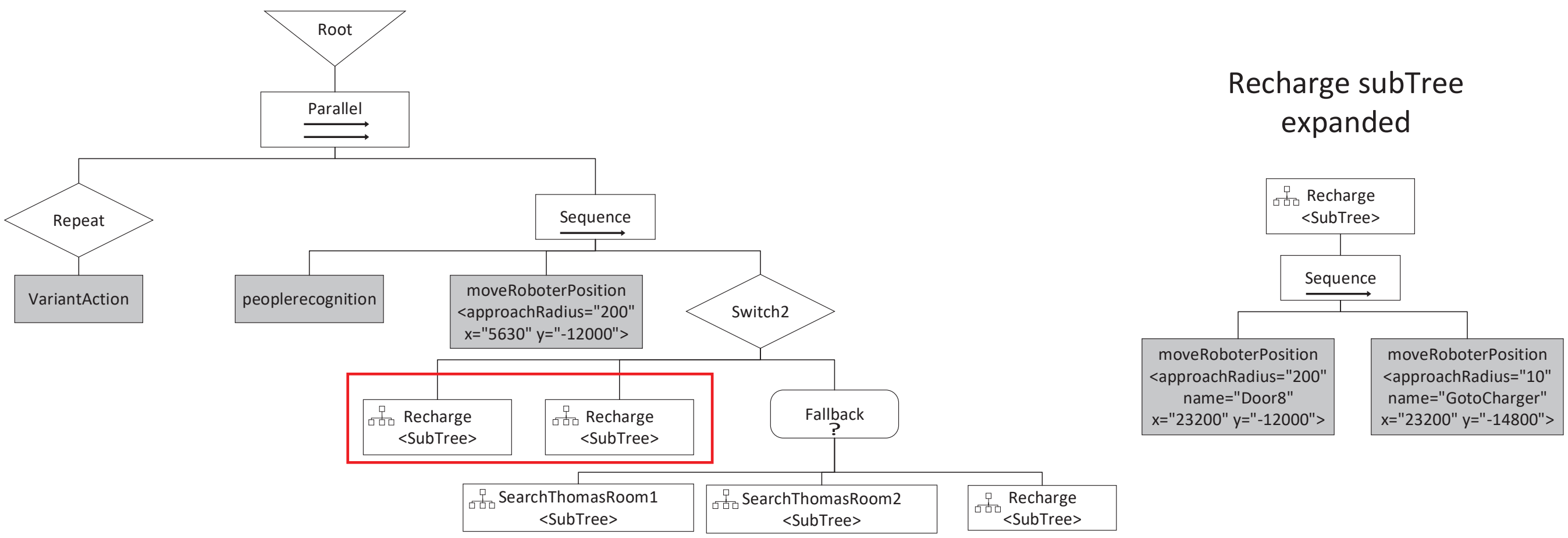

**Figure 4.** Behavior tree model of a retirement home robot from project bundles. The red box highlights an example of a reuse by reference for a sub-tree Recharge (expanded on the right side). A legend is shown in Figure 2.

*Reuse by clone-and-own* was used slightly less frequently than reuse by reference (in 48% of behavior tree models). In projects with multiple behavior tree models, we observe that, when two behavior trees have the same activities, the similar parts (a branch in the tree, a sub-tree or the entire model) are reused after some minor changes, such as adding new nodes or removing old ones. The Dyno project in Fig. 5, a drone-based parcel delivery project, includes two behavior tree models: one for a parcel delivery mission (M1) and another one for a route scheduler mission (M2). These models are an example of clone-and-own, where the developer reused the entire behavior tree model for two different missions that share similar activities after proper modification depending on the mission.[11]

*Reuse by reference through file inclusion* was used in 40% of the projects (10 of the 25 projects). Repeated activities were implemented as action nodes in header files that were passed later in the main tree execution file. This allows projects with multiple models having similar activities to reuse these actions in different parts of the model. Zooming into these projects, 30% of them belong to `BehaviorTree.CPP` projects, and 70% to `py_trees_ros` projects. Going back to our discussion in Sect. 4, these numbers can be related to `BehaviorTree.CPP` having a dedicated XML format to express the behavior tree model, so reuse by inclusion is done on the (C++) source-code level code, which we speculate can be challenging to use and maintain by the developers. While behavior tree models in `py_trees_ros` are intertwined with the Python code, making it easier to reuse by inclusion. However, the nature of the projects and the frequency of needed changes might have influenced the developer choice of reuse mechanism in both languages, since changing on the action level can be done easily without the challenge of going through every model in this mechanism. However, these conjectures need to be confirmed by feedback from the projects' developers, which we see as valuable future work.

> Observation 6. We conjecture that the identified simple reuse mechanisms suffice for the identified robotics projects. It is less clear whether it would be useful to have more safe and rich reuse mechanisms known from mainstream programming languages, including namespacing and safe reuse contracts (interfaces), which tend to be heavyweight for users to learn and use. More research is needed to determine whether sufficiently lightweight and safe reuse mechanism could be realized.

## 6 Threats to Validity

**Internal.** The major threat that could affect the results about the models are possible errors in our Python scripts calculating the model metrics. As a form of a quality check, we manually counted node types and checked the script results against these after building the model. We excluded commented parts and unused node types in the behavior trees codes.

When comparing behavior trees to other UML diagrams, we only conducted a comparison to behavior trees concepts and whether the UML diagrams support them or not. Thus, we might have missed other concepts offered by those two UML languages, but not behavior trees, which could have highlighted the limitation of Behavior trees. In a research extension, we plan to mitigate that.

**External.** The list of identified open source robotic projects might be missing examples from Bitbucket and GitLab. Both platforms are used in the robotics community, however, they do not provide a code search API, which made it difficult to conduct a code-level search. We conducted a less precise query in Bitbucket and GitLab using *behavior trees* as a search term in the web interface, however, we could not identify any real robotics projects from that search.

[11] The model can be found in full-size in the online appendix, [1] in addition to the models of the other projects.

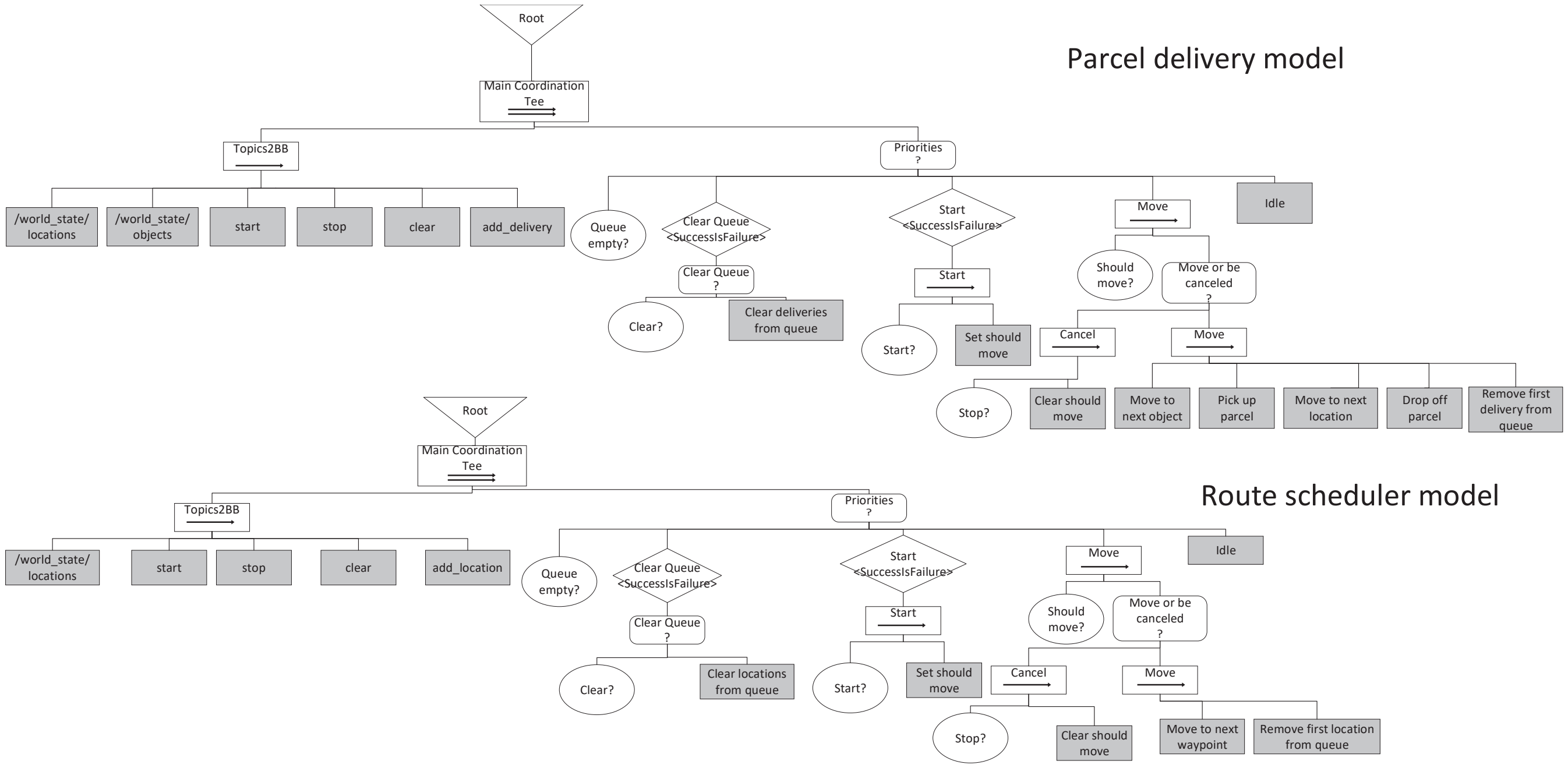

**Figure 5.** An example of clone-and-own referencing in Behavior trees from project Dyno. Each model belong to a different mission (M1) parcel delivery, and (M2) a route scheduler. Legend in Figure 2.

We have only considered projects using Python and C++ libraries with ROS support, while there might be other open-source robotics projects out there. We acknowledge that limiting our search to ROS-supported languages might have resulted in missing other robotic projects. However, we focused on the two dominant languages in ROS, assuming that this is the most representative framework for open source robotics.

## 7 Related Work

Guidelines how to apply behavior trees as well as important model properties relevant for multi-robotics systems have been discussed before, by Colledanchise et al. [11–15]. However, these works do not provide real-world robotic projects to support the claims related to the model properties of behavior trees modularity, flexibility, and reusability. In contrast, we conducted an empirical study of behavior tree characteristics in real-world robotic projects, and in comparison to those literature, we were only able to observe reusability through analyzing the studied behavior tree models. So, our work can be considered as complementary, confirming some of the declared claims about behavior trees. However, more research is needed to support the other claims.

The use of behavior trees in various robotics sub-domains has also been discussed before. Colledanchise and Ögren [14], in their thorough introduction to behavior trees, discuss model excerpts from industrial applications that the authors are aware of (e.g., by the truck manufacturer SCANIA). They also discuss the relationship of behavior trees to other behavior models (e.g., finite state machine and decision tree). A survey Iovino et al. [30] of 160 research papers, devoted to the development of behavior tree as a tool for AI in games and robotics, have highlighted how behavior trees have been used in different application areas. In comparison to our work, we focus on comparing behavior trees modeling concepts and design principles from a language perspective. In addition, we provide actual behavior tree models in a community dataset mined from open-source robotic projects. which non of the previous literature did, which can be used for further research.

Bagnell et al. [3] present a robotic system with perception, planning, and control, where task control is driven by behavior trees. The authors find that behavior trees easily describe complex manipulation tasks, and that behaviors can be reused. They chose behavior trees, because they had a team with a broad skill sets and needed a task orchestrating model that is easy to by each team member. Our findings support their claim to some extent, assuming all team members have basic programming skills. However, we noticed that behavior trees require language-oriented programming skills.

## 8 Conclusion

We presented a study of behavior trees languages and their use in real-world robotics applications. We systematically compared the concepts of popular behavior tree language implementations with each other and with two other established UML languages for describing behavior (state machines and activity diagrams). We mined open-source projects from code repositories and extracted their behavior tree models from the codebases, analyzing their characteristics and use of concepts. We contribute a dataset of models in the online appendix [1], together with scripts, and additional data.

Our analysis sheds light on languages designed outside of the language-engineering community for the vibrant and highly interesting domain of robotics. We believe that studying modeling and language-engineering practices is beneficial for both communities, as it helps to improve language-engineering methods and tools, as well as to improve the actual practices and languages. In fact, our results illustrate that many of the modeling and language-engineering methods are relevant in practice, especially the models-at-runtime paradigm but also reusability and meta-model extensibility. However, it also shows that developing languages in a rather pragmatic way, without hundreds of pages of specification documents and with a basic, but extensible meta-model, or even without an explicitly defined meta-model seems to be successful. Such a strategy seems to attract practitioners not trained in language and modeling technology, allowing practitioners who come from lower-level programming paradigms to raise the level of abstraction and effectively implement missions of robots in higher-level representations.

Still, we have observed aspects of behavior tree languages and models that are clearly suboptimal from the language design perspective, and pose interesting opportunities for this community to make impact. Behavior trees are a highly extensible language, but this comes at a cost of not having proper concrete syntax, and a seemingly high requirements that its users need to be familiar with language-oriented programming. Moreover, the abstract-syntax oriented modeling encourages heavy coupling of the model and the controlled system. This makes it really hard to work with models separately—for instance, verification, testing, and even visualizing may be a challenge without a working build system.

In the future, we would like to identify further sources of models and analyze them, as well as deepen the comparison with the traditional behavior-specification languages, which includes mining state machines represented in popular libraries (e.g., SMACH [6] or FlexBE [38]). Improving the syntax and semantics of behavior tree languages themselves is another interesting direction for future research we want to pursue.

## Acknowledgment

This work was partially supported by the Wallenberg AI, Autonomous Systems and Software Program (WASP) funded by the Knut and Alice Wallenberg Foundation, the EU H2020 project ROSIN (732287), and the SIDA project BRIGHT.